\documentclass[10pt,twocolumn,letterpaper]{article}

\usepackage[pagenumbers]{wacv}
\usepackage{colortbl}        % for \rowcolor
\usepackage{arydshln}        % for \cdashline
\usepackage{multirow}
\usepackage{tikz}
\usepackage{soul}
\usetikzlibrary{arrows.meta,positioning,decorations.pathreplacing,calc}

\newcommand{\ours}{\textsc{HyperSurg}}
\newcommand{\baseline}{SurgCLIP}
\newcommand{\bench}{\textsc{SurgHiBench}}

\newcommand{\RR}{\mathbb{R}}

\newcommand{\LL}{\mathbb{L}}
\DeclareMathOperator{\acosh}{acosh}
\DeclareMathOperator{\asin}{asin}

\definecolor{wacvblue}{rgb}{0.21,0.49,0.74}
\usepackage[pagebackref,breaklinks,colorlinks,allcolors=wacvblue]{hyperref}
\usepackage[capitalize,noabbrev]{cleveref}  % must be loaded after hyperref

\title{A Hierarchy-Aware Video-Language Model Evaluation and \\Hyperbolic Baseline for Surgery}

\author{
Ana Manzano Rodriguez$^{1,2,3,4}$ \quad
Pascal Mettes$^{1,2}$ \quad
Marlies P. Schijven$^{1,3,4,5}$ \quad
Cees G. M. Snoek$^{1,2}$\\[6pt]
{\small $^{1}$Data Science Center HAVA-Lab, University of Amsterdam}\\
{\small $^{2}$Video \& Image Sense Lab, Informatics Institute, University of Amsterdam}\\
{\small $^{3}$Amsterdam UMC Location University of Amsterdam, Surgery \quad
$^{4}$Amsterdam Public Health, Digital Health}\\
{\small $^{5}$Amsterdam Gastroenterology and Metabolism}
}

\begin{document}
\maketitle

%==============================================================================
\begin{abstract}
Surgical procedures follow a phase-to-step hierarchy, yet the video-language models used to recognize them are evaluated with flat per-level metrics that ignore cross-level coherence and error structure. In this paper we make two contributions to address this problem, (i) we introduce \bench{}, the first hierarchy-aware evaluation suite for surgical video understanding, with three tasks measuring recognition, consistency, and severity across granularity levels. We evaluate a general-purpose CLIP model, a Euclidean surgical model, and, as second contribution: (ii) \ours{}, a new hyperbolic model that enforces phase-step containment via entailment cones, across four (existing) datasets spanning three procedure types. The suite reveals that two models with the same accuracy can produce predictions of very different error severity, ranging from sibling confusions within the correct phase to unrelated cross-phase predictions. Hyperbolic geometry shifts predictions toward the correct procedural neighborhood, and these gains scale with the tree-likeness of each dataset's annotation hierarchy, providing a principled indicator when hierarchy-aware geometry helps.
\vspace{-2em}
\end{abstract}

%==============================================================================
\section{Introduction}
\label{sec:intro}

%---------- Figure 1: Hierarchy Motivation ----------
\begin{figure}[t]
  \centering
    \includegraphics[width=\linewidth]{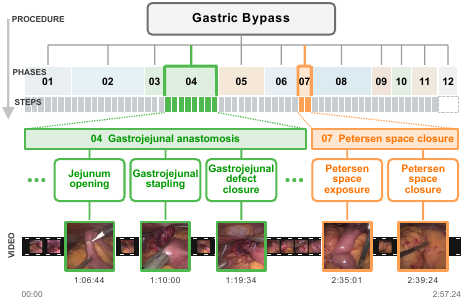}
    \vspace{-2em}
    \caption{\textbf{Surgical procedures are hierarchically structured.} A gastric bypass decomposes into 12 phases containing 46 steps. For surgical AI, current video-language models embed all levels in a flat space and evaluate each independently, ignoring underlying structure. To address this problem, we introduce (i) a hierarchy-aware evaluation suite and (ii) a hyperbolic baseline model. %Frames: StrasBypass70~\cite{lavanchy2024multibypass}, SBP52, 1 fps. 
    }
    \label{fig:hierarchy-page1}
    \vspace{-1em}
\end{figure}

Developing assistance systems for the operating room requires understanding what is happening during a procedure~\cite{maierhein2017surgical}. Surgical workflow analysis provides this understanding by recognizing surgical activities at different levels of temporal granularity, which is inherently hierarchical: a gallbladder removal (cholecystectomy) comprises \emph{phases} (\eg Calot triangle dissection), each containing \emph{steps} (\eg dissecting the cystic duct), which in turn consist of fine-grained \emph{actions} (\eg grasping, cutting) (\cref{fig:hierarchy-page1}). Because steps belong to broader procedural contexts, not all recognition errors are equal~\cite{bertinetto2020making,srivastava2023severity}. Confusing two steps within the same phase means the prediction is at least in the right part of the procedure, whereas predicting a step from an entirely different phase places it in an unrelated procedural context. Effective surgical support therefore requires recognizing both the broad phase and the specific step, and ensuring that these predictions are consistent.

Despite this clinical need, current surgical video-language models do not explicitly capture hierarchical relationships. State-of-the-art models such as SurgVLP~\cite{yuan2025surgvlp} learn from narrated surgical video without hierarchical structure, and while the SurgLaVi model~\cite{perez2025surglavi} learns from hierarchically annotated data, all levels still coexist in the model's flat embedding space without explicit hierarchical distinction. Evaluation has not addressed this gap either, as all current models are measured with accuracy or F1 at each level \textit{independently}, without assessing whether predictions across levels are consistent or whether errors respect the surgical procedure structure. In practice, two models with the same score may produce errors of very different severity, with standard metrics not being able to distinguish between them. This disconnect between what we evaluate and what clinical deployment demands has been identified as a key barrier for AI adoption in surgery~\cite{manzano2025bridging}.

To address the lack of hierarchy-awareness in surgical AI, this paper makes two core contributions: 
\begin{enumerate}
    \item \textbf{Contribution I: \bench{}} (\cref{sec:benchmark}), the first hierarchy-aware evaluation suite for surgical video-language models, with three tasks measuring per-level and joint recognition, cross-level consistency, and error severity, evaluated across four (existing) video datasets spanning three procedure types. \bench{} comes with three baseline models: CLIP, \baseline{}$_{(\beta)}^{*}$, and     
    \item \textbf{Contribution II: \ours{}} (\cref{sec:method}), a new hierarchy-aware surgical video-language baseline model that enforces entailment cone constraints over a hyperbolic embedding space,  where the exponentially growing volume naturally fits the tree-like structure of surgical taxonomies, paired with a hierarchy-aware triplet batch sampler.
\end{enumerate}
\bench{} reveals (\cref{sec:results}) that models with comparable accuracy differ substantially in how coherent their predictions are. With \ours{}, predictions shift from unrelated phases toward the correct procedural neighborhood, and cross-level contradictions decrease by up to 13\,pp. Step recognition benefits most from the hierarchical inductive bias (+9.3\,pp on average), as the exponentially growing space provides room for fine-grained distinctions that flat geometry collapses. 
Before detailing our contributions we first discuss related work.

%==============================================================================
\section{Related Work}
\label{sec:related}

\textbf{Surgical Vision-Language Models.} While vision-language models such as CLIP~\cite{radford2021clip} have transformed general visual recognition for quite a while, their adaptation to surgery is still in its early stages. SurgVLP~\cite{yuan2025surgvlp} was the first to apply contrastive pretraining to narrated surgical video, using a ResNet-50 vision encoder aligned with a BioClinicalBERT text encoder. 
SurgLaVi~\cite{perez2025surglavi} contributed a large-scale dataset of hierarchically annotated surgical clips spanning actions, steps, and phases, together with SurgCLIP, a dual-encoder model trained on this data. Subsequent work has aimed to make these representations hierarchy-aware: HecVL~\cite{yuan2024hecvl} constructs three separate embedding levels (clip, phase, video) via progressive aggregation of fine-grained representations into coarser ones, while PeskaVLP~\cite{yuan2024peskavlp} enriches textual descriptions with LLM-generated hierarchical knowledge and procedural alignment. These methods rely on increasingly complex architectures (separate embedding spaces, progressive aggregation, LLM-augmented knowledge) yet all remain in Euclidean space.
\ours{} takes an alternative approach: rather than adding architectural complexity, we change the underlying geometry, unifying all levels in a single hyperbolic space where the structure itself encodes hierarchical granularity.

\noindent\textbf{Hierarchical Evaluation.}
Not all misclassifications are equally costly: confusing adjacent categories in a semantic hierarchy is less severe than confusing distant ones, and evaluation metrics should reflect this.
Bertinetto \etal~\cite{bertinetto2020making} showed that classifiers trained to minimize tree distance to the ground truth sacrifice minimal flat accuracy while producing substantially milder errors. This principle has found clinical uptake.
In dermatology, hierarchy-aware prototypes penalize confusing melanoma with a benign nevus (a distant branch) more heavily than confusing it with basal cell carcinoma (a sibling malignancy), reducing dangerous misdiagnoses~\cite{yu2025skinlesion}.
In radiology, Srivastava and Mishra~\cite{srivastava2023severity} showed that models with comparable recognition ability on chest X-rays can differ substantially in error severity. Moreover, existing hierarchy-aware training losses and post-inference reranking methods, designed for taxonomies where only leaf nodes are labels, actually \emph{increase} severity on medical hierarchies where every node is a valid diagnosis. This suggests that reducing error severity in multi-level clinical hierarchies requires representations that intrinsically respect the hierarchy.
For action recognition, Long \etal~\cite{long2020searching} introduced sibling-mAP and cousin-mAP, which relax correctness criteria along a category tree.
However, all of these frameworks target natural-image classification or general action recognition; none has been applied to surgical video, where the procedural hierarchy provides an equally natural tree structure for grading error severity.
Within surgical workflow analysis, temporal hierarchy has been studied through phase recognition~\cite{czempiel2020tecno}, step recognition~\cite{lavanchy2024multibypass}, and instrument-action detection~\cite{ramesh2023dissecting}, yet these tasks are invariably evaluated independently with flat metrics.
Our cross-level consistency evaluation is, to our knowledge, the first to explicitly measure whether a model's predictions \emph{across} granularity levels are mutually consistent.

\noindent\textbf{Hyperbolic Representations.}
Hyperbolic spaces are non-Euclidean geometries of constant negative curvature whose volume grows exponentially with radius, making them a natural fit for encoding hierarchies~\cite{mettes2024hyperbolic}.  
Poincar\'e embeddings~\cite{nickel2017poincare} showed that hyperbolic spaces can embed taxonomies with orders-of-magnitude fewer dimensions than Euclidean ones, and Ganea \etal~\cite{ganea2018entailment} introduced a geometric notion of containment in hyperbolic space, where broader concepts enclose more specific ones within cone-shaped regions.
Building on these ideas, MERU~\cite{desai2023meru} was the first to apply hyperbolic geometry to vision-language pretraining, learning a two-level cross-modal hierarchy (text~$\supset$~image) on natural images.
HyCoCLIP~\cite{pal2025hycoclip} extended entailment to compositional sub-parts, and PHyCLIP~\cite{yoshikawa2026phyclip} proposed products of hyperbolic spaces for joint hierarchy and compositionality.
In medical imaging, HYDEN~\cite{qiao2025hyden} applied hyperbolic representations to radiology reports. 
We propose \ours{}, a hyperbolic surgical video-language model that enforces the procedural phase-step hierarchy through entailment cones. Concurrent to our work, HyperVLP by Hu \etal~\cite{hu2026hypervlp} also represents surgical video and language in a Lorentz manifold and use hyperbolic entailment to encode hierarchical relations. However, both HyperVLP and prior work evaluate with standard flat metrics, leaving open the hierarchy-aware questions our benchmark addresses.

%==============================================================================
\section{Hierarchy-Aware Evaluation for Surgery}
\label{sec:benchmark}

We introduce \bench{}, the first hierarchy-aware evaluation suite for surgical video-language models. \bench{} combines three complementary tasks that together provide a complete picture of hierarchical awareness: whether predictions at each granularity level are correct, whether they are mutually consistent along the levels, and whether predictions land on related or unrelated surgical concepts (\cref{sec:tasks}). These tasks are evaluated across four (existing) surgical video datasets that span a spectrum of tree-likeness in their annotation structure (\cref{sec:benchmarks}) and three models ranging from general-purpose to hierarchy-aware (\cref{sec:models}). \bench{} is model-agnostic and can evaluate any surgical video-language model that produces phase and step predictions. Each task is defined below together with its metrics, all computed per-video and averaged. All data, training details and source code will be made available on our benchmark website.

\subsection{Hierarchy-Aware Tasks}
\label{sec:tasks}
\noindent\textbf{Task 1: Single- and Multi-Level Recognition.}
\textit{How well does a model recognize phases and steps, individually and jointly?} We first evaluate how well a model identifies phases and steps independently, and report video-averaged \emph{accuracy} and \emph{macro-averaged F1} at each level. While accuracy measures overall correctness, macro-F1 better captures performance on rare but clinically important classes by weighting all classes equally, which is important given the severe class imbalance typical of surgical datasets. A model may achieve high phase accuracy while performing poorly on steps, or vice versa; to detect such imbalances we also report a
  \emph{multi-level} score, defined as the geometric mean of phase and step accuracy:
  \begin{equation}
      \text{Multi-level} = \sqrt{\text{Acc}_{\text{phase}} \times \text{Acc}_{\text{step}}},
      \label{eq:gm}
  \end{equation}
which remains low unless both levels are strong (\eg 30\% phase with 10\% step yields 17.3\%, while 20\% on both yields 20\%).
These per-level metrics do not show, however, whether predictions are coherent across levels, which the remaining tasks address.

%---------- Table: Evaluation Datasets ----------
\begin{table*}[t!]
\centering
\caption{\textbf{\bench{} evaluates video-language models on hierarchy-awareness}, covering three tasks and four surgical video datasets. Datasets are ordered by tree-likeness,  defined as the percentage of steps that belong to exactly one phase (100\%\,=\,pure tree, lower\,=\,more DAG-like). 
}
\label{tab:datasets}
\vspace{-0.5em}
\setlength{\tabcolsep}{5pt}
\small
\begin{tabular}{@{}lllrrrlr@{}}
\toprule
\textbf{Dataset} & \textbf{Source} & \textbf{Procedure} & \textbf{Videos} & \textbf{Phases}\textsuperscript{$\dagger$} & \textbf{Steps}\textsuperscript{$\dagger$} & \textbf{Prompts} & \textbf{Tree-likeness} \\
\midrule
\textit{MIPO} & Graeff \etal~\cite{graeff2025mipo} & Radius fracture & 50 & 5 & 8 & Ours & 100\% \\
\textit{StrasbyPass70} & Lavanchy \etal~\cite{lavanchy2024multibypass} & Gastric bypass & 70 & 11 & 46 & SurgLaVi \cite{perez2025surglavi} & 74\% \\
\textit{BernBypass70} & Lavanchy \etal~\cite{lavanchy2024multibypass} & Gastric bypass & 70 & 11 & 46 & SurgLaVi \cite{perez2025surglavi} & 54\% \\
\textit{GraSP} & Ayobi \etal~\cite{ayobi2025grasp} & Prostatectomy & 13 & 11 & 21 & SurgLaVi \cite{perez2025surglavi} & 33\% \\
\bottomrule
\multicolumn{8}{@{}l}{\textsuperscript{$\dagger$}\footnotesize Classes used for evaluation; see supplementary \cref{app:prompts} for details.}\\ 
\end{tabular}
\vspace{-0.5em}
\end{table*}

\noindent\textbf{Task 2: Cross-Level Consistency.}
\textit{Are the model's phase and step predictions mutually consistent?}
A system that predicts the \emph{Preparation} phase alongside a \emph{Clipping} step produces contradictory outputs that could mislead clinical decision support~\cite{way2003causes}.
We measure joint correctness across both levels with three metrics. The \emph{Both-wrong} and \emph{Both-right} rates measure how often both levels are simultaneously wrong or right, revealing whether errors are isolated to one level or propagate across both. \emph{Step\textbar Phase} is step accuracy conditioned on the phase being correct, which isolates step discrimination from phase errors.
These metrics are inspired by hierarchical consistency measures in image classification~\cite{jiang2025hcal,park2025hcast}, adapted to surgical video, where only joint phase-step correctness has been reported~\cite{ramesh2021multitask} without analyzing error patterns across levels.

\noindent\textbf{Task 3: Hierarchy-Aware Error Severity.}
\textit{Does the model confuse related or unrelated surgical concepts?}
We capture this with two complementary measures. \emph{Sibling step accuracy}~\cite{long2020searching} measures how often the predicted step falls within the same phase as the ground-truth step, even if the exact step is wrong. \emph{Cousin step accuracy} measures the opposite: how often the prediction lands in an entirely different phase. A model with high sibling and low cousin accuracy predicts within the correct procedural context, indicating that it captures the coarse structure of the procedure. We also introduce \emph{semantic step accuracy}, which measures whether the model recognizes the type of surgical action being performed regardless of anatomical site (\eg \emph{Jejunojejunal Defect Closure} \vs \emph{Gastrojejunal Defect Closure}), identifying \emph{what} is being done even when it confuses \emph{where}. We manually group steps that share the same surgical action across different anatomical sites; full groupings are in the supplemental (\cref{app:semantic}).

\subsection{Surgical Video Datasets}
\label{sec:benchmarks}
\noindent\textbf{Evaluation Datasets.} Our evaluation focuses on phase and step recognition, the two granularity levels that enable the cross-level analysis required by our three tasks. We evaluate on four video datasets (\cref{tab:datasets}): MIPO~\cite{graeff2025mipo} (50 radius fracture fixation videos), StrasbyPass70 and BernBypass70~\cite{lavanchy2024multibypass} (70 gastric bypass videos each, recorded at two institutions), and GraSP~\cite{ayobi2025grasp} (13 robot-assisted prostatectomy videos). All four provide phase and step annotations, but differ in how steps relate to phases. In some datasets, each step belongs to exactly one phase, forming a clean tree (MIPO, 100\%). In others, steps recur across multiple phases, forming a directed acyclic graph (DAG) where the parent--child relationship is ambiguous (GraSP, 33\%). This difference stems from how each dataset defines its steps. GraSP's step vocabulary mixes procedural substages (\emph{Seminal vesicle dissection}, \emph{Prevessical dissection}) with reusable instrument actions (\emph{Suction}, \emph{Cut suture or tissue}, \emph{Tie suture}). The latter are phase-agnostic---\emph{Suction} occurs whenever there is blood or fluid (10 of 11 phases), \emph{Cut suture or tissue} whenever tissue must be separated (10 of 11). By contrast, the bypass datasets (54--74\%) mostly define steps as procedural substages (\emph{Gastrojejunal Defect Closure}, \emph{Retrogastric Dissection}, \emph{Vertical Stapling}) that describe the surgical goal being accomplished, and are therefore largely phase-exclusive by construction. MIPO follows the same goal-level convention (\emph{Distal fixation}, \emph{Proximal fixation}). 
This spectrum allows our evaluation to also test whether the benefits of hierarchy-aware representations are tied to the structural properties of each dataset's annotation hierarchy.

For zero-shot evaluation, we use the text descriptions provided by Perez~\etal~\cite{perez2025surglavi} for GraSP and the bypass datasets; for MIPO, we wrote new prompts following the same narration style (see supplemental \cref{app:prompts}).

\noindent\textbf{Pretraining Dataset.} 
The surgical video-language models evaluated in this work are pretrained on a subset of the SurgLaVi dataset~\cite{perez2025surglavi}. The full dataset contains 5{,}317 narrated surgical videos sourced from diverse surgical domains, across specialties such as general surgery, colorectal, gynecology, thoracic, and urology. Of these, 2{,}464 are part of the open-source subset, called SurgLaVi-$\beta$. We use SurgLaVi-$\beta$ without modification.

Each video is divided into temporal segments at three hierarchy levels, with captions generated from the narrator's descriptions, yielding ${\sim}$113K caption--clip pairs. These levels differ in temporal granularity and scope: \emph{Actions} (Level~1) capture brief, action-focused surgical moments (\eg ``the surgeon uses a stapler to make the first transverse staple line''). \emph{Steps} (Level~2) aggregate multiple actions into intermediate procedural substages with more context (\eg ``the surgeon dissects into the lesser sac; the first transverse stapler firing is performed''). \emph{Phases} (Level~3) span longer temporal segments with rich procedural detail (\eg ``the surgeon begins the gastric bypass by placing trocars at the angle of His, dissecting along the lesser curvature with a perigastric technique''). 
This hierarchical structure mirrors the procedural hierarchy of surgery itself, making hierarchy-aware training possible and allowing the model to learn representations that respect the natural granularity of surgical procedures.

\subsection{Baseline Models}
\label{sec:models}
Our evaluation suite comes with three models. Two existing models and one model that we propose as part of this paper.
\textbf{(i)}~\textit{CLIP ViT-B/16}~\cite{radford2021clip}, the original OpenAI CLIP model with no surgical pretraining. Since CLIP has no temporal modeling, each frame is encoded independently and similarities are averaged over the temporal window. It serves as a lower bound on what general-purpose vision-language representations achieve on surgical tasks. 
\textbf{(ii)}~\textit{\baseline{}$_{(\beta)}^{*}$} is our reimplementation of SurgCLIP$_{(\beta)}$ by Perez \etal~\cite{perez2025surglavi} with identical architecture (TimeSformer~\cite{bertasius2021timesformer} + BERT-base~\cite{devlin2019bert}), training procedure, and Euclidean cosine similarity.
Since SurgLaVi released model and evaluation code but not the training pipeline, we reimplemented training from the details in their paper; the reproduction closely matches their reported results across all benchmarks (see \cref{tab:downstream_full} in supplemental), confirming a faithful reimplementation. We follow their zero-shot evaluation protocol and test splits, adapting the scripts for Lorentz distance in \ours{}.
\textbf{(iii)}~\ours{} is our hierarchy-aware model operating on a Lorentz hyperboloid, described next, using the same architecture, training data, and number of parameters as \baseline{}$_{(\beta)}^{*}$, ensuring a controlled experiment where any difference with our model is attributable to the embedding geometry alone. 

%----------------

%==============================================================================
\section{Hyperbolic Baseline Model for Surgery} 
\label{sec:method}
Having defined how to evaluate hierarchy-awareness, we now propose a baseline model that encodes surgical hierarchy by operating in hyperbolic space, where the temporal granularity of procedures is naturally represented. We cover the geometric background (\cref{sec:lorentz}), the architecture (\cref{sec:architecture}), and the training objectives (\cref{sec:losses}).

\subsection{Background: Lorentz Hyperbolic Space}
\label{sec:lorentz}
Standard contrastive models, like CLIP and SurgCLIP, measure similarity via cosine distance, which depends only on the angle between embeddings and discards their norms.
This leaves a single degree of freedom---angular direction---to simultaneously separate semantically distinct videos and distinguish hierarchy levels within the same video, competing demands that one signal cannot satisfy.
The Lorentz hyperboloid decouples these two roles: the distance from the origin encodes generality while the angular direction independently encodes semantic content, providing the missing degree of freedom.
Formally, it is defined as the upper part of a two-sheeted hyperboloid in $(n{+}1)$-dimensional Minkowski space with curvature $-1/c$:
\begin{equation}
    \LL^{n}_{c} = \{\mathbf{x} \in \RR^{n+1} : \langle \mathbf{x}, \mathbf{x} \rangle_{\LL} = -1/c,\; x_0 > 0\},
\end{equation}
where $\langle \mathbf{x}, \mathbf{y} \rangle_{\LL} = -x_0 y_0 + \sum_{i=1}^{n} x_i y_i$ is the Lorentz inner product.
The geodesic distance between two points is:
\begin{equation}
    d_{\LL}(\mathbf{x}, \mathbf{y}) = \frac{1}{\sqrt{c}}\,\acosh\bigl(-c\,\langle \mathbf{x}, \mathbf{y} \rangle_{\LL}\bigr).
    \label{eq:lorentz_dist}
\end{equation}

\begin{figure}[t]
  \centering
  \includegraphics[width=0.9\linewidth]{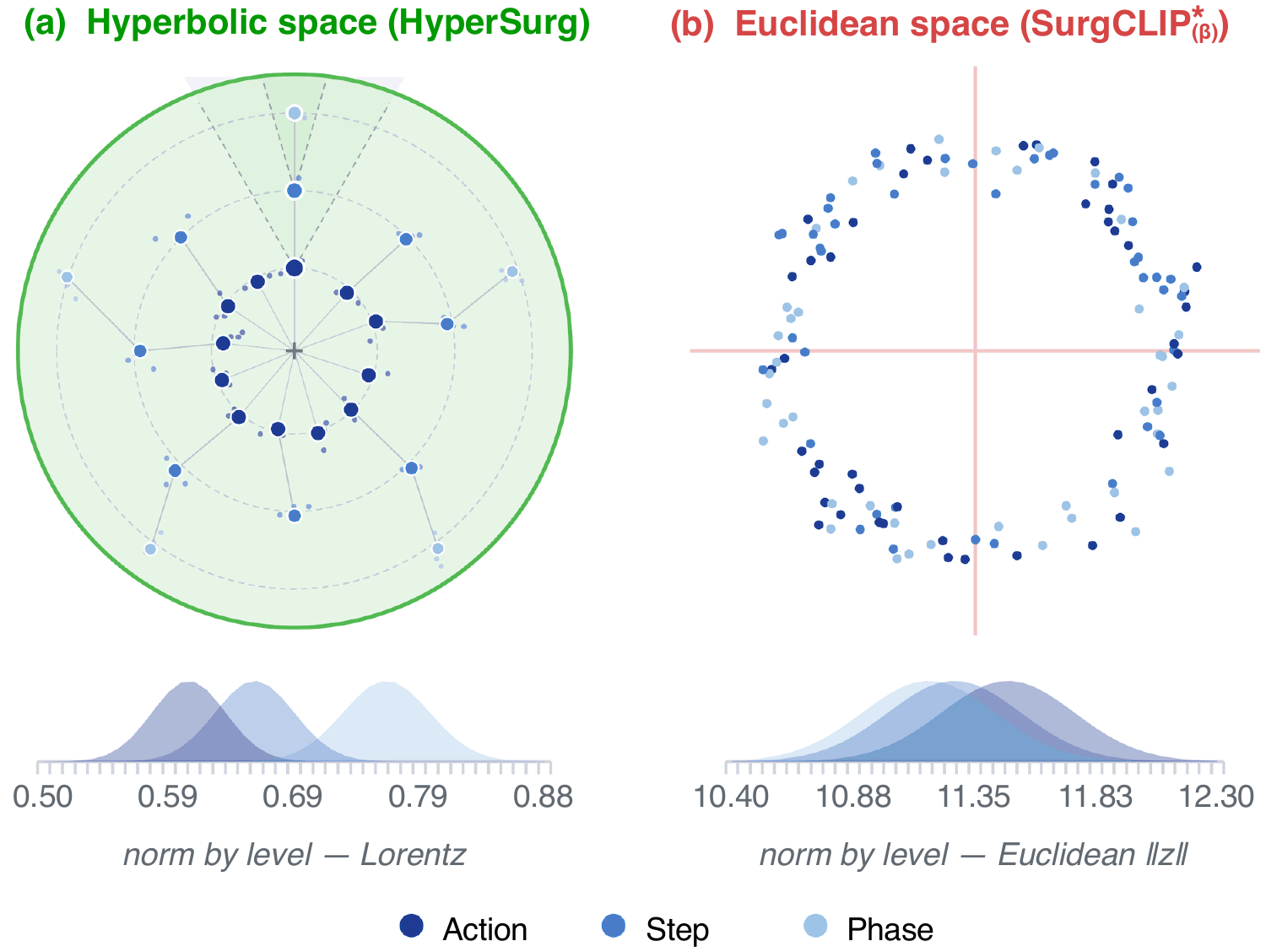}
\caption{\textbf{Hyperbolic geometry organizes the surgical hierarchy by embedding norm; Euclidean cosine space collapses it.}
  \textbf{(a)}~In \ours{} (green), embedding norms increase from fine-grained actions to coarse phases, reflecting the hierarchical structure. Gray wedges are entailment cones (\cref{eq:aperture}).
  \textbf{(b)}~In \baseline{}$_{(\beta)}^{*}$ (red), L2 normalization for cosine similarity discards any norm signal, preventing cosine-based models from exploiting the hierarchy.
  \textbf{Bottom:} per-level norm distributions. Note distinguishability of hierarchical levels in hyperbolic space.
  }
  \label{fig:embeddings}
\end{figure}

A key property is that the volume of a ball of radius $r$ grows with ratio $e^{(n-1)r/\sqrt{c}}$, whereas in Euclidean space the same volume grows only polynomially ($\propto r^n$).
This exponential capacity makes the radial axis a natural encoding of hierarchy, separating granularity levels that flat geometry collapses. \Cref{fig:embeddings} illustrates this property: in \ours{}, embedding norms increase monotonically from fine-grained actions to coarse phases, whereas L2 normalization in cosine-based models collapses this separation entirely. 

%---------- Figure 4: Embedding Visualization ----------

\subsection{Architecture}
\label{sec:architecture}

%---------- Figure 2: Architecture (Dual Encoder Pipeline) ----------
\begin{figure*}[t]
    \centering
    \begin{tikzpicture}
        \node[inner sep=0pt, rounded corners=5pt, clip] {\includegraphics[width=\textwidth]{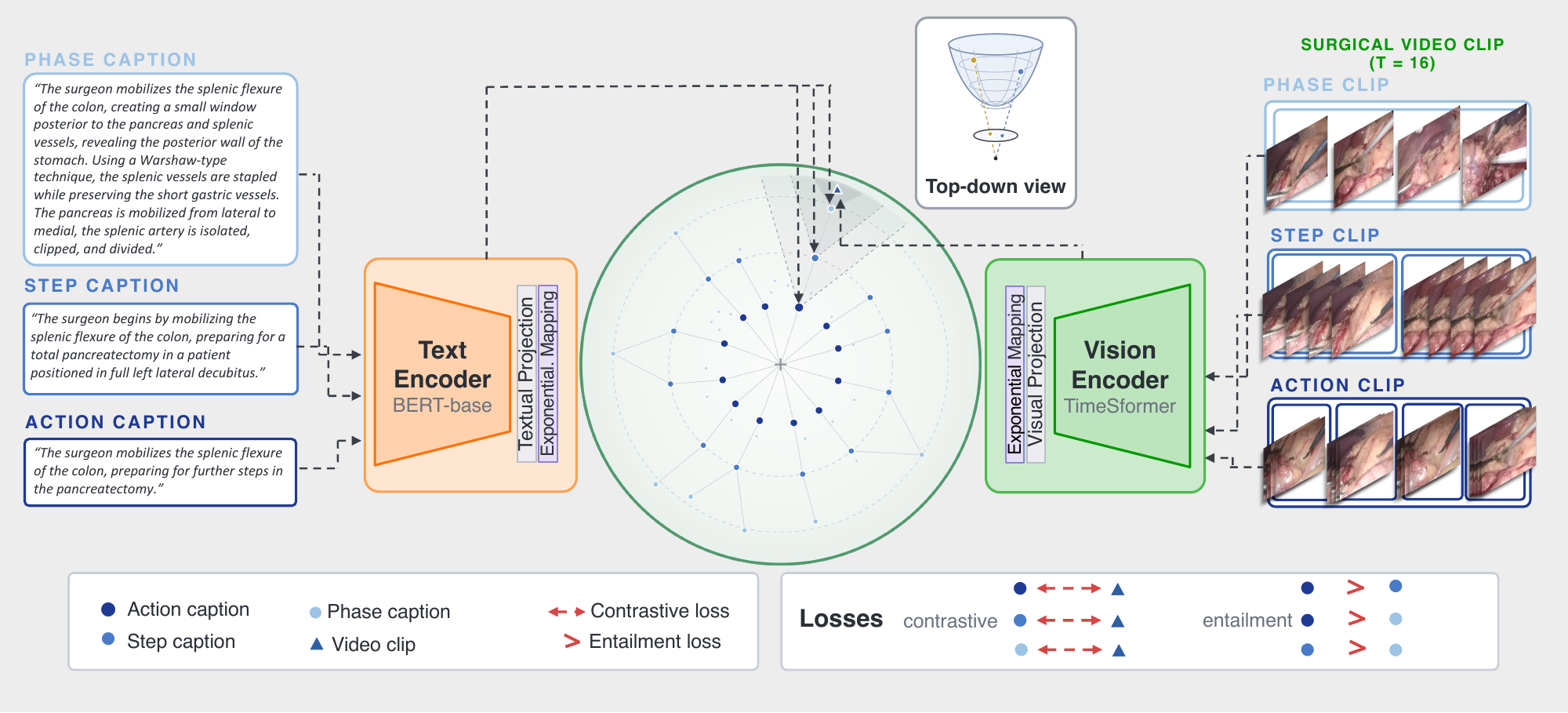}};
    \end{tikzpicture}
    \caption{\textbf{Overview of the proposed hierarchy-aware \ours{} baseline model.} Surgical video clips at three hierarchy levels %\cs{(fine/mid/coarse)}
    (action, step, phase) are encoded by a TimeSformer (ViT-B/16), while the corresponding narration-derived captions are encoded by BERT-base. Both branches project into a shared 256-D Lorentz hyperboloid $\LL^{256}_{c}$ via learned projections and the exponential map. The contrastive loss $\mathcal{L}_{\text{contr}}$ aligns video--text pairs via Lorentz distance (\cref{eq:sim}); the entailment loss $\mathcal{L}_{\text{entail}}$ enforces parent--child containment across hierarchy levels within the same cone (\cref{eq:entail}). 
    }
    \label{fig:architecture}
\end{figure*}

\ours{} follows a dual-encoder paradigm where a video encoder and a text
encoder are jointly trained to align surgical clips with hierarchical
captions in a shared hyperbolic embedding space (\cref{fig:architecture}).
To isolate the effect of geometry, \ours{} uses the same encoders as
\baseline{}$_{(\beta)}^{*}$: a
TimeSformer~\cite{bertasius2021timesformer} (ViT-B/16, divided
space-time attention, 12 layers) encodes $T{=}16$ uniformly sampled
frames into a 768-dimensional video representation via mean-pooled
spatial patch tokens, and a BERT-base~\cite{devlin2019bert}
text encoder maps captions to 768-dimensional text representations via
the [CLS] token.

The two architectures diverge after the encoders. Both project the 768-dimensional representations to $\RR^{256}$ via learned linear layers, but \baseline{}$_{(\beta)}^{*}$ then applies L2-normalization to measure cosine similarity, while \ours{} scales them by learnable per-modality log-space factors $\alpha_v, \alpha_t$ and maps them onto the Lorentz hyperboloid $\LL^{256}_{c}$ via the exponential map at the origin
$\mathbf{o} = (1/\!\sqrt{c},\, \mathbf{0})$:
\begin{equation}
    \exp_{\mathbf{o}}^{c}(\mathbf{v}) =
    \cosh\!\bigl(\sqrt{c}\,\|\mathbf{v}\|_{\LL}\bigr)\,\mathbf{o} \;+\;
    \frac{\sinh\!\bigl(\sqrt{c}\,\|\mathbf{v}\|_{\LL}\bigr)}
    {\sqrt{c}\,\|\mathbf{v}\|_{\LL}}\,\mathbf{v},
    \label{eq:expmap}
\end{equation}
where $\mathbf{v} \in T_{\mathbf{o}}\LL^{n}_{c}$ is the tangent vector at $\mathbf{o}$. Only the space components are stored; the time component is recovered via the hyperboloid constraint $x_0 = \sqrt{1/c + \|\mathbf{x}_{\text{space}}\|^2}$ when needed for distance computations. The curvature $c$ is a learnable parameter stored in log-space and clamped to $[0.1, 10]$.

\subsection{Training Objectives}
\label{sec:losses}

\noindent\textbf{Lorentz Contrastive Loss.}
We measure similarity via negative Lorentz distance scaled by a learnable temperature $\tau$:
\begin{equation}
    s(\mathbf{v}_i, \mathbf{t}_j) = -\,d_{\LL}(\mathbf{v}_i, \mathbf{t}_j)\,/\,\tau.
    \label{eq:sim}
\end{equation}
The loss follows symmetric InfoNCE over all GPU-gathered pairs:
\begin{align}
    \mathcal{L}_{\text{contr}} &= \frac{1}{2}\bigl(\mathcal{L}_{\text{v2t}} + \mathcal{L}_{\text{t2v}}\bigr), \nonumber\\
    \mathcal{L}_{\text{v2t}} &= -\frac{1}{B}\sum_{i}\log\frac{e^{s(\mathbf{v}_i,\mathbf{t}_i)}}{\sum_{j} e^{s(\mathbf{v}_i,\mathbf{t}_j)}}.
    \label{eq:contrastive}
\end{align}

\noindent\textbf{Hierarchy Entailment Loss.}
The three-level hierarchy described in \cref{sec:benchmarks} forms a tree.
Following Ganea \etal~\cite{ganea2018entailment}, we enforce this containment via entailment cones.
For a parent text embedding $\mathbf{p}$, its entailment cone has half-aperture:
\begin{equation}
    \psi(\mathbf{p}) = \asin\!\left(\frac{2\,r_{\min}}{\|\mathbf{p}\|_{\RR}\,\sqrt{c}}\right),
    \label{eq:aperture}
\end{equation}
where $r_{\min}$ controls the minimum cone width.
A child $\mathbf{q}$ (\eg a step or phase caption from the same video) must lie within the parent's cone.
Let $\phi(\mathbf{p}, \mathbf{q})$ denote the exterior angle at $\mathbf{p}$ in the hyperbolic triangle $O$--$\mathbf{p}$--$\mathbf{q}$ (where $O$ is the origin).
The entailment loss penalizes violations with a slack factor $\eta$:
\begin{equation}
    \mathcal{L}_{\text{entail}} = \frac{1}{|\mathcal{P}|}\sum_{(\mathbf{p},\mathbf{q})\in\mathcal{P}} \max\!\bigl(0,\;\phi(\mathbf{p},\mathbf{q}) - \eta\,\psi(\mathbf{p})\bigr),
    \label{eq:entail}
\end{equation}
where $\mathcal{P}$ is the set of parent--child pairs within each batch and $\eta{=}1.2$ provides geometric slack.

\noindent\textbf{Hierarchical Triplet Sampling.}
During training, each batch consists of video-caption pairs sampled from the dataset. Standard random sampling rarely produces within-video hierarchical pairs: with captions spread across thousands of videos, a random batch of a few hundred samples is unlikely to contain multiple hierarchy levels from the same video, leaving the entailment loss effectively without training signal.
We introduce a novel hybrid sampler that composes each batch of $B$ samples from $B{-}3k$ random samples and $k{=}3$ forced \emph{triplets}, each containing one caption per hierarchy level from the same video.
This guarantees $\sim$9 entailment pairs per batch while maintaining contrastive diversity.

\noindent\textbf{Total Loss.}
\begin{equation}
    \mathcal{L} = \mathcal{L}_{\text{contr}} + \lambda\,\mathcal{L}_{\text{entail}},
    \label{eq:total}
\end{equation}
with $\lambda{=}0.1$, selected from a range of values (\cref{tab:lambda}). An ablation study confirms that both the entailment cones and the hierarchy-aware triplet batch sampler contribute beyond Lorentz distance alone (\cref{app:ablation}).

\noindent\textbf{Training details.} We train for 50 epochs on 8$\times$ NVIDIA RTX A5000 GPUs with an effective batch size of 312, using AdamW ($\text{lr}{=}10^{-4}$, weight decay$\,{=}\,0.02$) with OneCycleLR (5-epoch warmup, cosine annealing). Video frames are augmented with random resized crop (scale 0.5--1.0) and horizontal flip. Since no training split is provided with the dataset, we hold out 15\% of the videos (369 out of 2{,}464) for monitoring training. Both \ours{} and \baseline{}$_{(\beta)}^{*}$ are trained with the same procedure, data, and hyperparameters, including masking same-video captions from the contrastive denominator to prevent conflicting gradients from clips that share a source video. 
Full details are provided in supplemental \cref{app:training}.

%==============================================================================
\section{Results}
\label{sec:results}

\subsection{Task 1: Single- and Multi-Level Recognition}
\label{sec:results_multilevel}

%---------- Table 1: Phase + Step + GM on multi-level datasets ----------
\begin{table}[t]
\centering
\caption{\textbf{Task 1: Single- and Multi-Level Recognition.} Zero-shot video-averaged accuracy (\%), \textbf{bold}\,=\,best per dataset. Overall phase and step recognition both improve with hierarchy-awareness, though per-dataset performance varies. The multi-level score, which requires both levels to be strong, tends to benefit from hierarchy-awareness, with gains scaling with tree-likeness. 
}
\label{tab:multilevel}
\setlength{\tabcolsep}{2.5pt}
\small
\begin{tabular}{@{}lrr@{\hskip 8pt}rr@{\hskip 8pt}r@{}}
\toprule
& \multicolumn{2}{c}{\textbf{Phase ($\uparrow$)}} & \multicolumn{2}{c}{\textbf{Step ($\uparrow$)}} & \multirow{2}{*}{\textbf{Multi-level ($\uparrow$)}} \\
\cmidrule(lr){2-3}\cmidrule(lr){4-5}
& \textit{Accuracy} & \textit{F1} & \textit{Accuracy} & \textit{F1} & \\
\midrule
\multicolumn{6}{@{}>{\columncolor{gray!15}[0pt][0pt]}l@{}}{\textit{MIPO}} \\
CLIP & 16.2 & 6.6 & 17.0 & 4.9 & \cellcolor{cyan!4} 16.6 \\
\baseline{}$_{(\beta)}^{*}$ & 13.6 & 7.9 & 8.3 & 5.6 & \cellcolor{cyan!4} 10.6 \\
\ours{} & \textbf{20.9} & \textbf{16.2} & \textbf{17.6} & \textbf{8.9} & \cellcolor{cyan!4} \textbf{19.2} \\
\multicolumn{6}{@{}>{\columncolor{gray!15}[0pt][0pt]}l@{}}{\textit{StrasbyPass70}} \\
CLIP & 18.5 & 4.2 & 3.3 & 0.7 & \cellcolor{cyan!4} 7.8 \\
\baseline{}$_{(\beta)}^{*}$ & 29.5 & 22.7 & 17.2 & 7.4 & \cellcolor{cyan!4} 22.5 \\
\ours{} & \textbf{30.2} & \textbf{25.2} & \textbf{32.0} & \textbf{12.0} & \cellcolor{cyan!4} \textbf{31.1} \\
\multicolumn{6}{@{}>{\columncolor{gray!15}[0pt][0pt]}l@{}}{\textit{BernBypass70}} \\
CLIP & 20.9 & 4.2 & 1.8 & 0.3 & \cellcolor{cyan!4} 6.1 \\
\baseline{}$_{(\beta)}^{*}$ & \textbf{23.9} & \textbf{17.4} & 13.5 & 5.6 & \cellcolor{cyan!4} 18.0 \\
\ours{} & 17.4 & 12.5 & \textbf{29.7} & \textbf{8.5} & \cellcolor{cyan!4} \textbf{22.7} \\
\multicolumn{6}{@{}>{\columncolor{gray!15}[0pt][0pt]}l@{}}{\textit{GraSP}} \\
CLIP & 4.3 & 1.5 & 6.4 & 1.2 & \cellcolor{cyan!4} 5.2 \\
\baseline{}$_{(\beta)}^{*}$ & 25.1 & \textbf{17.7} & \textbf{17.9} & \textbf{9.0} & \cellcolor{cyan!4} \textbf{21.2} \\
\ours{} & \textbf{26.6} & 17.0 & 14.7 & 7.8 & \cellcolor{cyan!4} 19.8 \\
\midrule
\multicolumn{6}{@{}>{\columncolor{cyan!4}[0pt][0pt]}l@{}}{\textbf{\textit{Overall}}} \\
CLIP & 15.0 & 4.1 & 7.1 & 1.8 & \cellcolor{cyan!4} 8.9 \\
\baseline{}$_{(\beta)}^{*}$ & 23.0 & 16.4 & 14.2 & 6.9 & \cellcolor{cyan!4} 18.1 \\
\ours{} & \textbf{23.8} & \textbf{17.7} & \textbf{23.5} & \textbf{9.3} & \cellcolor{cyan!4} \textbf{23.2} \\
\bottomrule
\end{tabular}
\end{table}

We present results for Task 1 in \Cref{tab:multilevel},  where we compare
recognition performance across granularity levels. The CLIP baseline, which has no surgical pretraining, provides a cautionary note about interpreting accuracy in isolation. On MIPO, CLIP achieves higher accuracy than the Euclidean surgical model at both levels (16.2\% vs 13.6\% phase, 17.0\% vs 8.3\% step), yet its F1 is lower in both cases (6.6 vs 7.9, 4.9 vs 5.6), indicating that CLIP collapses predictions into the most frequent classes, inflating accuracy while failing to distinguish rarer categories. The pattern holds at the aggregate level, where CLIP's overall F1 remains near chance (4.1 phase, 1.8 step).

Surgical pretraining in Euclidean space substantially improves phase recognition (23.0\% accuracy, F1 16.4), but step accuracy lags behind (14.2\%). These results indicate that coarse-grained surgical categories are already separable in flat Euclidean space, while steps require finer distinctions among classes that share a common parent phase, and this is where hierarchical structure in the embedding geometry makes the difference. With \ours{}, step accuracy rises to 23.5\% while phase accuracy remains comparable (23.8\%), narrowing the gap between the two levels. The multi-level score, which requires both levels to be strong, confirms that the benefit comes from making phase and step recognition work together rather than improving either in isolation (overall 23.2\% vs 18.1\%), though per-dataset gains depend on the degree of tree-structure in each annotation hierarchy. The gains track this spectrum closely: the bypass datasets (54--74\% tree-likeness) show strong multi-level improvements, and MIPO (100\% tree-likeness) sees the largest. The exception is GraSP, where \baseline{}$_{(\beta)}^{*}$ leads on step accuracy: its mixed-granularity step vocabulary means that 67\% of steps recur across multiple phases (\cref{sec:benchmarks}), breaking the parent--child containment that entailment cones enforce. The more tree-like the hierarchy, the larger the gains, exactly the expected behaviour for a geometric inductive bias designed to exploit tree structure. 

Worth noting, MIPO is an orthopaedic procedure outside the surgical domains covered by the pretraining data~\cite{perez2025surglavi}, yet \ours{} achieves the best accuracy and F1 at both levels, suggesting that the hierarchical inductive bias generalises to out-of-distribution procedures where domain-specific pretraining alone falls short.

%---------- Table 2: Cross-level consistency ----------
\begin{table}[t]
\centering
\caption{\textbf{Task 2: Cross-Level Consistency} (\%). \textbf{Bold}\,=\,best per dataset.
As models become more hierarchy-aware, catastrophic failures where both levels are wrong decrease, and a correct phase prediction more often leads to a correct step, showing that the two granularity levels become coupled rather than independent.  
}

\label{tab:consistency}
\setlength{\tabcolsep}{3pt}
\small
\begin{tabular}{@{}lrrr@{}}
\toprule
 & \textbf{Both Wrong}\,($\downarrow$) & \textbf{Both Right}\,($\uparrow$) & \textbf{Step \textbar ~Phase}\,($\uparrow$) \\
\midrule
\multicolumn{4}{@{}>{\columncolor{gray!15}[0pt][0pt]}l@{}}{\textit{MIPO}} \\
CLIP         & 82.2 & \textbf{15.4} & \textbf{94.3} \\
\baseline{}$_{(\beta)}^{*}$ & 82.1 & 4.0  & 27.2 \\
\ours{}      & \textbf{70.9} & 9.4  & 44.3 \\
\multicolumn{4}{@{}>{\columncolor{gray!15}[0pt][0pt]}l@{}}{\textit{StrasbyPass70}} \\
CLIP         & 80.4 & 2.2  & 10.9 \\
\baseline{}$_{(\beta)}^{*}$ & 57.9 & 4.5  & 15.2 \\
\ours{}      & \textbf{44.8} & \textbf{7.0}  & \textbf{22.8} \\
\multicolumn{4}{@{}>{\columncolor{gray!15}[0pt][0pt]}l@{}}{\textit{BernBypass70}} \\
CLIP         & 78.1 & 0.8  & 3.5 \\
\baseline{}$_{(\beta)}^{*}$ & 66.1 & 3.5  & 14.5 \\
\ours{}      & \textbf{57.2} & \textbf{4.3}  & \textbf{24.2} \\
\multicolumn{4}{@{}>{\columncolor{gray!15}[0pt][0pt]}l@{}}{\textit{GraSP}} \\
CLIP         & 89.9 & 0.7  & 20.9 \\
\baseline{}$_{(\beta)}^{*}$ & \textbf{67.3} & \textbf{10.2} & \textbf{35.3} \\
\ours{}      & 67.8 & 9.1  & 30.9 \\
\midrule
\multicolumn{4}{@{}>{\columncolor{cyan!4}[0pt][0pt]}l@{}}{\textbf{\textit{Overall}}} \\
CLIP         & 82.6 & 4.8  & \textbf{32.4} \\
\baseline{}$_{(\beta)}^{*}$ & 68.3 & 5.6  & 23.0 \\
\ours{}      & \textbf{60.2} & \textbf{7.4}  & 30.6 \\
\bottomrule
\end{tabular}
\end{table}

\subsection{Task 2: Cross-Level Consistency}
\label{sec:results_consistency}

We present results for Task~2 in \Cref{tab:consistency}, where we evaluate whether phase and step predictions are mutually coherent. The Both Wrong rates are high across the board, which shows how challenging zero-shot recognition at two granularity levels is. CLIP fails on the vast majority of frames (82.6\% Both Wrong), and appears to lead on Both Right and Step\textbar Phase on MIPO, but this is the same class-collapse artefact from Task~1. With a phase F1 of only 6.6, Step\textbar Phase is computed on very few frames and the high score is misleading. These metrics should therefore be read jointly with Task~1, as they can overstate coherence when overall accuracy is low.

The consistent reduction in Both Wrong from CLIP (82.6\%) to \baseline{}$_{(\beta)}^{*}$ (68.3\%) to \ours{} (60.2\%) shows that surgical pretraining helps and encoding hierarchical structure aids further, with the largest drop on StrasbyPass70 (from 57.9\% for \baseline{}$_{(\beta)}^{*}$ to 44.8\% for \ours{}). The exception is again GraSP, where \baseline{}$_{(\beta)}^{*}$ leads for the same structural reasons discussed in Task~1. Step\textbar Phase shows that when the phase is correct, the hierarchy-aware model picks the right step more often (30.6\% vs 23.0\% overall), suggesting that hierarchy-awareness couples the two levels through the embedding geometry.

%---------- Table 3: Hierarchy-aware error severity ----------
\begin{table}[t]
\centering
\caption{\textbf{Task 3: Error severity}  (\%). 
\textbf{Bold}\,=\,best per dataset. 
The hierarchy-aware model's predictions more often fall within the correct phase and preserve the surgical action, indicating that its step predictions are more context-aware.}
\label{tab:hier_acc}
\setlength{\tabcolsep}{3pt}
\small
\begin{tabular}{@{}lrrr@{}}
\toprule
 & \textbf{Sibling}\,($\uparrow$) & \textbf{Cousin}\,($\downarrow$) & \textbf{Semantic Step}\,($\uparrow$) \\
\midrule
\multicolumn{4}{@{}>{\columncolor{gray!15}[0pt][0pt]}l@{}}{\textit{MIPO}} \\
CLIP         & 20.6 & 79.4 & 20.4 \\
\baseline{}$_{(\beta)}^{*}$ & 21.9 & 78.1 & 9.1  \\
\ours{}      & \textbf{27.9} & \textbf{72.1} & \textbf{24.2} \\
\multicolumn{4}{@{}>{\columncolor{gray!15}[0pt][0pt]}l@{}}{\textit{StrasbyPass70}} \\
CLIP         & 20.2 & 79.8 & 8.8 \\
\baseline{}$_{(\beta)}^{*}$ & 43.4 & 56.6 & 30.9 \\
\ours{}      & \textbf{51.3} & \textbf{48.7} & \textbf{44.6} \\
\multicolumn{4}{@{}>{\columncolor{gray!15}[0pt][0pt]}l@{}}{\textit{BernBypass70}} \\
CLIP         & 21.3 & 78.7 & 9.9 \\
\baseline{}$_{(\beta)}^{*}$ & 36.8 & 63.2 & 20.6 \\
\ours{}      & \textbf{46.3} & \textbf{53.7} & \textbf{40.9} \\
\multicolumn{4}{@{}>{\columncolor{gray!15}[0pt][0pt]}l@{}}{\textit{GraSP}} \\
CLIP         & 14.0 & 86.0 & 10.7 \\
\baseline{}$_{(\beta)}^{*}$ & \textbf{31.6} & \textbf{68.4} & \textbf{29.1} \\
\ours{}      & 29.2 & 70.8 & 24.4 \\
\midrule
\multicolumn{4}{@{}>{\columncolor{cyan!4}[0pt][0pt]}l@{}}{\textbf{\textit{Overall}}} \\
CLIP         & 19.0 & 81.0 & 12.4 \\
\baseline{}$_{(\beta)}^{*}$ & 33.4 & 66.6 & 22.4 \\
\ours{}      & \textbf{38.7} & \textbf{61.3} & \textbf{33.5} \\
\bottomrule
\end{tabular}
\end{table}

\subsection{Task 3: Hierarchy-Aware Error Severity}
\label{sec:results_severity}
\begin{figure}[t]
  \centering
  \includegraphics[width=\linewidth]{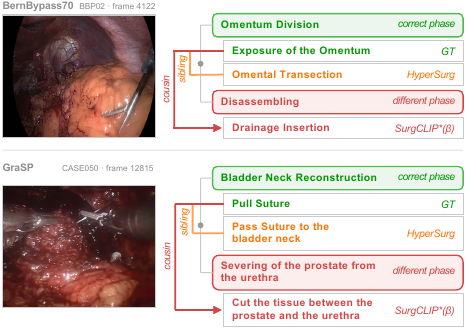}
\caption{\textbf{Per-level metrics cannot distinguish between errors.} On both frames, each model predicts the correct phase but the wrong step, so accuracy and F1 record the same outcome. The errors differ in where they land in the hierarchy: the {\color{orange!90!red}orange} prediction is a sibling of the ground truth, a different step within the correct phase, while the {\color{red!75}red} one is a cousin, from an unrelated phase. \bench{} measures this distinction. 
}
\label{fig:qual}
\end{figure}

The previous tasks measured \emph{whether} the model is correct; this task examines \emph{where} predictions fall in the procedural hierarchy and whether they preserve the surgical action. We present results in \Cref{tab:hier_acc}. CLIP shows the lowest sibling accuracy across all datasets (19.0\% overall), indicating that its predictions fall in unrelated phases almost uniformly. Surgical pretraining improves this substantially (33.4\% sibling accuracy), meaning a larger share of step predictions fall within the correct phase, and hierarchy-awareness pushes it further (38.7\%). Semantic step accuracy follows the same trend, rising from 12.4\% (CLIP) to 22.4\% (\baseline{}$_{(\beta)}^{*}$) to 33.5\% (\ours{}), meaning the model more often recognises the surgical action being performed even when it confuses the anatomical site. The top confusions illustrate what these numbers mean in practice. On BernBypass70, \ours{}'s dominant confusion for 
\emph{Gastrojejunal Anastomosis Reinforcement} is \emph{Gastrojejunal Defect Closure}, a sibling step within the same phase that involves suturing the same anatomical structure. By contrast, \baseline{}$_{(\beta)}^{*}$ confuses it with \emph{Horizontal Stapling}, a step from a different phase (Gastric Pouch Creation) at a different anatomical site with a different surgical goal. \Cref{fig:qual} shows the same pattern on GraSP and BernBypass70: the hierarchy-aware prediction remains within the correct phase, the Euclidean one does not. Flat accuracy treats both confusions identically, but these results provide insights that only hierarchy-aware evaluation can report and that directly address the needs of the surgical domain, where models with similar accuracy can produce errors of very different severity that should not be overlooked. 

We provide qualitative examples of successful and failed predictions for both \baseline{}$_{(\beta)}^{*}$ and \ours{}, across all four datasets, in supplementary \cref{fig:temporal_crosslevel_1}.

%==============================================================================
\section{Conclusion}
\label{sec:conclusion} 
In this work, we posit that phases and steps in surgical video should not be treated as independent recognition tasks. We introduce a hierarchy-aware evaluation suite that measures cross-level consistency and error severity alongside per-level accuracy, showing across four datasets and three procedure types that video-language models with comparable accuracy can result in very different error patterns. We propose \ours{}, a hierarchy-aware baseline that produces more accurate and consistent cross-level predictions by aligning the embedding geometry with the hierarchical structure of procedures. 
The resulting gains scale with the tree-likeness of each dataset's annotation hierarchy, providing a principled indicator of when hierarchy-aware representations will help. As richer annotations become available, we expect both the evaluation framework and the geometric approach to extend naturally to deeper hierarchies, providing the community with tools to build surgical workflow systems that are not only accurate, but evaluated in ways that reflect what matters in the operating room.
%==============================================================================
\section*{Acknowledgements}
This work was supported by the University of Amsterdam's Data Science Centre, as part of the HAVA Lab.

%==============================================================================
% References
 {\small 
  \bibliographystyle{ieeenat_fullname}
  \bibliography{references}                

@article{ayobi2025grasp,
  author    = {Nicolas Ayobi and Santiago Rodriguez and Alejandra Perez and Isabela Hernandez and Nicolas Aparicio and Eugenie Dessevres and Sebastian Pe{\~n}a and Jessica Santander and Juan Ignacio Caicedo and Nicolas Fernandez and Pablo Arbelaez},
  title     = {Pixel-Wise Recognition for Holistic Surgical Scene Understanding},
  journal   = {Medical Image Analysis},
  pages     = {103726},
  year      = {2025},
}

@inproceedings{bertasius2021timesformer,
  author    = {Gedas Bertasius and Heng Wang and Lorenzo Torresani},
  title     = {Is Space-Time Attention All You Need for Video Understanding?},
  booktitle = {International Conference on Machine Learning (ICML)},
  year      = {2021},
}

@inproceedings{bertinetto2020making,
  author    = {Luca Bertinetto and Romain Mueller and Konstantinos Tertikas and Sina Samangooei and Nicholas A. Lord},
  title     = {Making Better Mistakes: Leveraging Class Hierarchies with Deep Networks},
  booktitle = {Proceedings of the IEEE/CVF Conference on Computer Vision and Pattern Recognition (CVPR)},
  pages     = {12503--12512},
  year      = {2020},
}

@inproceedings{czempiel2020tecno,
  author    = {Tobias Czempiel and Magdalini Paschali and Matthias Keicher and Walter Simson and Hubertus Feussner and Seong Tae Kim and Nassir Navab},
  title     = {{TeCNO}: Surgical Phase Recognition with Multi-Stage Temporal Convolutional Networks},
  booktitle = {International Conference on Medical Image Computing and Computer-Assisted Intervention (MICCAI)},
  pages     = {343--352},
  year      = {2020},
}

@inproceedings{desai2023meru,
  author    = {Karan Desai and Maximilian Nickel and Tanmay Rajpurohit and Justin Johnson and Ramakrishna Vedantam},
  title     = {Hyperbolic Image-Text Representations},
  booktitle = {International Conference on Machine Learning (ICML)},
  pages     = {7694--7731},
  year      = {2023},
}

@inproceedings{devlin2019bert,
  author    = {Jacob Devlin and Ming-Wei Chang and Kenton Lee and Kristina Toutanova},
  title     = {{BERT}: Pre-Training of Deep Bidirectional Transformers for Language Understanding},
  booktitle = {North American chapter of the association for computational linguistics: human language technologies (NAACL-HLT)},
  pages     = {4171--4186},
  year      = {2019},
}

@inproceedings{ganea2018entailment,
  author    = {Octavian-Eugen Ganea and Gary B{\'e}cigneul and Thomas Hofmann},
  title     = {Hyperbolic Entailment Cones for Learning Hierarchical Embeddings},
  booktitle = {International Conference on Machine Learning (ICML)},
  pages     = {1646--1655},
  year      = {2018},
}

@article{graeff2025mipo,
  author    = {Camille Gra{\"e}ff and Nicolas Padoy and Philippe Liverneaux and Thomas Lampert},
  title     = {Introducing Surgical Workflow Recognition in Orthopaedic Surgery with Timestamp Supervision},
  journal   = {Computers in Biology and Medicine},
  volume    = {197},
  pages     = {110995},
  year      = {2025},
}

@article{hu2026hypervlp,
  author    = {Yaojun Hu and Kun Yuan and Nassir Navab and Haochao Ying and Jian Wu and Nicolas Padoy},
  title     = {{HyperVLP}: Enhancing Hierarchical Surgical Video-Language Pre-Training in Hyperbolic Space},
  journal   = {arXiv preprint arXiv:2606.31245},
  year      = {2026},
}

@inproceedings{jiang2025hcal,
  author    = {Ruobing Jiang and Mengzhe Liu and Haobing Liu and Yanwei Yu},
  title     = {Hierarchy-Consistent Learning and Adaptive Loss Balancing for Hierarchical Multi-Label Classification},
  booktitle = {ACM International Conference on Information and Knowledge Management (CIKM)},
  year      = {2025},
}

@article{lavanchy2024multibypass,
  author    = {Joel L. Lavanchy and Sanat Ramesh and Diego Dall'Alba and Cristians Gonzalez and Paolo Fiorini and Beat P. M{\"u}ller-Stich and Philipp C. Nett and Jacques Marescaux and Didier Mutter and Nicolas Padoy},
  title     = {Challenges in Multi-Centric Generalization: Phase and Step Recognition in {Roux-en-Y} Gastric Bypass Surgery},
  journal   = {International Journal of Computer Assisted Radiology and Surgery},
  volume    = {19},
  pages     = {2249--2257},
  year      = {2024},
}

@inproceedings{long2020searching,
  author    = {Teng Long and Pascal Mettes and Heng Tao Shen and Cees G. M. Snoek},
  title     = {Searching for Actions on the Hyperbole},
  booktitle = {Proceedings of the IEEE/CVF Conference on Computer Vision and Pattern Recognition (CVPR)},
  pages     = {1138--1147},
  year      = {2020},
}

@article{maierhein2017surgical,
  author    = {Lena Maier-Hein and Swaroop S. Vedula and Stefanie Speidel and Nassir Navab and Ron Kikinis and Adrian Park and Matthias Eisenmann and Hubertus Feussner and Germain Forestier and Stamatia Giannarou and Makoto Hashizume and Darko Katic and Hannes Kenngott and Michael Kranzfelder and Anand Malpani and Keno M{\"a}rz and Thomas Neumuth and Nicolas Padoy and Carla Pugh and Nicolai Schoch and Danail Stoyanov and Russell Taylor and Martin Wagner and Gregory D. Hager and Pierre Jannin},
  title     = {Surgical Data Science for Next-Generation Interventions},
  journal   = {Nature Biomedical Engineering},
  volume    = {1},
  number    = {9},
  pages     = {691--696},
  year      = {2017},
}

@article{manzano2025bridging,
  author    = {Ana Manzano Rodriguez and Cees G. M. Snoek and Marlies P. Schijven},
  title     = {Bridging the Gap: Exposing the Hidden Challenges Towards Adoption of Artificial Intelligence in Surgery},
  journal   = {British Journal of Surgery},
  volume    = {112},
  number    = {11},
  year      = {2025},
}

@article{mettes2024hyperbolic,
  author    = {Pascal Mettes and Mina Ghadimi Atigh and Martin Keller-Ressel and Jeffrey Gu and Serena Yeung},
  title     = {Hyperbolic Deep Learning in Computer Vision: A Survey},
  journal   = {International Journal of Computer Vision},
  volume    = {132},
  number    = {9},
  pages     = {3484--3508},
  year      = {2024},
}

@inproceedings{nickel2017poincare,
  author    = {Maximilian Nickel and Douwe Kiela},
  title     = {Poincar{\'e} Embeddings for Learning Hierarchical Representations},
  booktitle = {Advances in Neural Information Processing Systems (NeurIPS)},
  pages     = {6338--6347},
  year      = {2017},
}

@inproceedings{pal2025hycoclip,
  author    = {Avik Pal and Max van Spengler and Guido Maria D'Amely di Melendugno and Alessandro Flaborea and Fabio Galasso and Pascal Mettes},
  title     = {Compositional Entailment Learning for Hyperbolic Vision-Language Models},
  booktitle = {International Conference on Learning Representations (ICLR)},
  year      = {2025},
}

@inproceedings{park2025hcast,
  author    = {Seulki Park and Youren Zhang and Stella X. Yu and Sara Beery and Jonathan Huang},
  title     = {Visually Consistent Hierarchical Image Classification},
  booktitle = {International Conference on Learning Representations (ICLR)},
  year      = {2025},
}

@article{perez2025surglavi,
  author    = {Alejandra Perez and Chinedu Nwoye and Ramtin Raji Kermani and Omid Mohareri and Muhammad Abdullah Jamal},
  title     = {Surg{L}a{V}i: Large-Scale Hierarchical Dataset for Surgical Vision-Language Representation Learning},
  journal   = {Medical Image Analysis},
  pages     = {103982},
  volume = {110},
  year      = {2026},
}

@inproceedings{qiao2025hyden,
  author    = {Zhi Qiao and Linbin Han and Xiantong Zhen and Jiahong Gao and Zhen Qian},
  title     = {{HYDEN}: Hyperbolic Density Representations for Medical Images and Reports},
  booktitle = {International Conference on Computational Linguistics (COLING)},
  year      = {2025},
}

@inproceedings{radford2021clip,
  author    = {Alec Radford and Jong Wook Kim and Chris Hallacy and Aditya Ramesh and Gabriel Goh and Sandhini Agarwal and Girish Sastry and Amanda Askell and Pamela Mishkin and Jack Clark and Gretchen Krueger and Ilya Sutskever},
  title     = {Learning Transferable Visual Models from Natural Language Supervision},
  booktitle = {International Conference on Machine Learning (ICML)},
  pages     = {8748--8763},
  year      = {2021},
}

@article{ramesh2021multitask,
  author    = {Sanat Ramesh and Diego Dall'Alba and Cristians Gonzalez and Tong Yu and Pietro Mascagni and Didier Mutter and Jacques Marescaux and Paolo Fiorini and Nicolas Padoy},
  title     = {Multi-Task Temporal Convolutional Networks for Joint Recognition of Surgical Phases and Steps in Gastric Bypass Procedures},
  journal   = {International Journal of Computer Assisted Radiology and Surgery},
  volume    = {16},
  pages     = {1111--1119},
  year      = {2021},
}

@article{ramesh2023dissecting,
  author    = {Sanat Ramesh and Vinkle Srivastav and Deepak Alapatt and Tong Yu and Aditya Murali and Luca Sestini and Chinedu Innocent Nwoye and Idris Hamoud and Saurav Sharma and Antoine Fleurentin and Georgios Exarchakis and Alexandros Karargyris and Nicolas Padoy},
  title     = {Dissecting Self-Supervised Learning Methods for Surgical Computer Vision},
  journal   = {Medical Image Analysis},
  volume    = {88},
  pages     = {102844},
  year      = {2023},
}

@article{srivastava2023severity,
  author    = {Satwik Srivastava and Deepak Mishra},
  title     = {Severity of Error in Hierarchical Datasets},
  journal   = {Scientific Reports},
  volume    = {13},
  pages     = {21903},
  year      = {2023},
}

@article{twinanda2017endonet,
  author    = {Andru P. Twinanda and Sherif Shehata and Didier Mutter and Jacques Marescaux and Michel De Mathelin and Nicolas Padoy},
  title     = {Endo{N}et: A Deep Architecture for Recognition Tasks on Laparoscopic Videos},
  journal   = {IEEE Transactions on Medical Imaging},
  volume    = {36},
  number    = {1},
  pages     = {86--97},
  year      = {2017},
}

@article{wagner2023heichole,
  author    = {Martin Wagner and Beat-Peter M{\"u}ller-Stich and Anna Kisilenko and Duc Tran and Patrick Heger and Lars M{\"u}ndermann and David M. Lubotsky and others},
  title     = {Comparative Validation of Machine Learning Algorithms for Surgical Workflow and Skill Analysis with the {HeiChole} Benchmark},
  journal   = {Medical Image Analysis},
  volume    = {86},
  pages     = {102770},
  year      = {2023},
}

@inproceedings{wang2022autolaparo,
  author    = {Ziyi Wang and Bo Lu and Yonghao Long and Fangxun Zhong and Tak-Hong Cheung and Qi Dou and Yunhui Liu},
  title     = {{AutoLaparo}: A New Dataset of Integrated Multi-Tasks for Image-Guided Surgical Automation in Laparoscopic Hysterectomy},
  booktitle = {International Conference on Medical Image Computing and Computer-Assisted Intervention (MICCAI)},
  pages = {486--496},
  year      = {2022},
}

@article{way2003causes,
  author    = {Lawrence W. Way and Lygia Stewart and Walter Gantert and Kingsway Liu and Crystine M. Lee and Karen Whang and John G. Hunter},
  title     = {Causes and Prevention of Laparoscopic Bile Duct Injuries: Analysis of 252 Cases from a Human Factors and Cognitive Psychology Perspective},
  journal   = {Annals of Surgery},
  volume    = {237},
  number    = {4},
  pages     = {460--469},
  year      = {2003},
}

@inproceedings{yoshikawa2026phyclip,
  author    = {Daiki Yoshikawa and Takashi Matsubara},
  title     = {{PHyCLIP}: $\ell_1$-Product of Hyperbolic Factors Unifies Hierarchy and Compositionality in Vision-Language Representation Learning},
  booktitle = {International Conference on Learning Representations (ICLR)},
  year      = {2026},
}

@article{yu2025skinlesion,
  author    = {Zhen Yu and Toan D. Nguyen and Lie Ju and Yaniv Gal and Maithili Sashindranath and Paul Bonnington and Lei Zhang and Victoria Mar and Zongyuan Ge},
  title     = {Hierarchical Skin Lesion Image Classification with Prototypical Decision Tree},
  journal   = {npj Digital Medicine},
  volume    = {8},
  pages     = {26},
  year      = {2025},
}

@inproceedings{yuan2024hecvl,
  author    = {Kun Yuan and Vinkle Srivastav and Nassir Navab and Nicolas Padoy},
  title     = {{HecVL}: Hierarchical Video-Language Pretraining for Zero-Shot Surgical Phase Recognition},
  booktitle = {International Conference on Medical Image Computing and Computer-Assisted Intervention (MICCAI)},
  year      = {2024},
}

@inproceedings{yuan2024peskavlp,
  author    = {Kun Yuan and Vinkle Srivastav and Nassir Navab and Nicolas Padoy},
  title     = {Procedure-Aware Surgical Video-Language Pretraining with Hierarchical Knowledge Augmentation},
  booktitle = {Advances in Neural Information Processing Systems (NeurIPS)},
  year      = {2024},
}

@article{yuan2025surgvlp,
  author    = {Kun Yuan and Vinkle Srivastav and Tong Yu and Joel L. Lavanchy and Jacques Marescaux and Pietro Mascagni and Nassir Navab and Nicolas Padoy},
  title     = {Learning Multi-Modal Representations by Watching Hundreds of Surgical Video Lectures},
  journal   = {Medical Image Analysis},
  volume    = {105},
  pages     = {103644},
  year      = {2025},
}
  }

%==============================================================================
% Supplementary Material

\appendix 
\maketitlesupplementary
\section{\bench{} Evaluation Suite}
\subsection{Semantic Step Groupings}
\label{app:semantic}

These groupings are used to compute the semantic step accuracy reported in Task~3 (\cref{sec:results_severity}). Two steps are ``semantically related'' if they perform the same type of action at different anatomical sites or are closely related surgical variants. The groupings were defined manually by inspecting each dataset's step vocabulary and identifying steps that share the same surgical action. \Cref{tab:semantic_groups} lists all groupings per dataset.

\begin{table}[th!]
\centering
\caption{\textbf{Semantic step groupings} used for Task~3. Steps within each group share the same surgical action at different anatomical sites.}
\label{tab:semantic_groups}
\setlength{\tabcolsep}{3pt}
\small
\begin{tabular}{@{}l p{5.2cm}@{}}
\toprule
\textbf{Group} & \textbf{Steps} \\
\midrule
\multicolumn{2}{@{}>{\columncolor{gray!15}[0pt][0pt]}l@{}}{\textit{MIPO (8 steps)}} \\
Fixation        & Distal fixation, Proximal fixation \\
Plate handling  & Introduction of the plate, Modification of plate positioning \\
\midrule
\multicolumn{2}{@{}>{\columncolor{gray!15}[0pt][0pt]}l@{}}{\textit{StrasbyPass70 / BernBypass70 (46 steps)}} \\
Defect closure     & Gastrojejunal Defect Closure, Jejunojejunal Defect Closure \\
Reinforcement      & Gastrojejunal Anast.\ Reinforcement, Jejunojejunal Anast.\ Reinforcement \\
Stapling           & Horizontal, Vertical, Gastrojejunal, Jejunojejunal Stapling \\
Limb measurement   & Biliary Limb Measurement, Alimentary Limb Measurement \\
Limb opening       & Jejunum Opening, Biliary Limb Opening, Alimentary Limb Opening \\
Dissection         & Fat Pad, Lesser Curvature, Retrogastric Dissection \\
Space closure/exposure & Petersen Space Exposure/Closure, Mesenteric Defect Exposure/Closure \\
\midrule
\multicolumn{2}{@{}>{\columncolor{gray!15}[0pt][0pt]}l@{}}{\textit{GraSP (21 steps)}} \\
Lymph node dissection   & Dissection Illiac Lymph Nodes, Dissection Obturator Lymph Nodes \\
Anatomical dissection   & Prevessical Dissection, Prostate Dissection, Seminal Vessicle Dissection, Denon Dissection \\
Packing            & Pack Lymph Nodes, Pack Prostate \\
Cutting            & Cut Prostate, Cut, Cut Bladder \\
Pass suture        & Pass Suture Urethra, Pass Suture Neck \\
\bottomrule
\end{tabular}
\end{table}

\subsection{Zero-Shot Evaluation and MIPO Prompts}
\label{app:prompts}
All zero-shot evaluations use text descriptions as prompts, with no prompt engineering, ensembling, or templates (e.g.\ ``a photo of \{\}'').
The model computes cosine similarity (or Lorentz distance for \ours{}) between the video embedding and each text prompt to produce per-class scores. For GraSP, BernBypass70, and StrasbyPass70, we use the text descriptions provided by SurgLaVi~\cite{perez2025surglavi}, not from the original dataset annotation files. BernBypass70 and StrasbyPass70 share the same annotation vocabulary and prompts; the dataset defines 12 phases~\cite{lavanchy2024multibypass}, but following SurgLaVi~\cite{perez2025surglavi} and existing zero-shot studies~\cite{yuan2024peskavlp}, we exclude the ``other'' class for fair comparison. MIPO is not part of SurgLaVi and its annotations do not include text descriptions, so we wrote prompts following the same narration style, describing the surgical actions visible in each phase and step. \Cref{tab:prompts_mipo_phase,tab:prompts_mipo_step} list the MIPO prompts.
%---------- MIPO Phases ----------
\begin{table*}[th!]
\centering
\caption{\textbf{Zero-shot prompts: MIPO phases} (5 classes). The MIPO repository documentation reports 4 phases and 6 steps; the released annotation files contain 5 and 8 distinct labels, respectively. We use the annotation file counts throughout.}
\label{tab:prompts_mipo_phase}
\small
\setlength{\tabcolsep}{4pt}
\begin{tabular}{@{}p{1.5cm} p{14.8cm}@{}}
\toprule
Phase & Prompt \\
\midrule
Approach & I incise the skin over the flexor carpi radialis tendon and elevate the pronator quadratus to expose the distal radius. \\
Closure & I close the incision with intradermal sutures and apply a sterile compress bandage over the wound. \\
Fixation & I place the volar locking plate and secure the distal radius fracture with K-wires, distal locking screws and proximal screws. \\
Installation & I prepare the volar locking plate with the aiming guide, apply a Velpeau bandage and inflate the pneumatic tourniquet. \\
Verification & I perform fluoroscopic controls with anteroposterior and lateral views and check flexor tendon freedom and wrist mobility. \\
\bottomrule
\end{tabular}
\end{table*}

%---------- MIPO Steps ----------
\begin{table*}[th!]
\centering
\caption{\textbf{Zero-shot prompts: MIPO steps} (8 classes; see \cref{tab:prompts_mipo_phase} caption).}
\label{tab:prompts_mipo_step}
\small
\setlength{\tabcolsep}{4pt}
\begin{tabular}{@{}p{3.5cm} p{12.8cm}@{}}
\toprule
Step & Prompt \\
\midrule
Approach & I incise the skin over the flexor carpi radialis tendon and elevate the pronator quadratus muscle to expose the fracture. \\
Closure & I close the wound with intradermal absorbable sutures and apply a sterile compress dressing. \\
Distal Fixation & I place K-wires and insert distal locking screws near the wrist joint using a drill guide. \\
Installation & I prepare the volar locking plate with the aiming guide and inflate the pneumatic tourniquet. \\
Introduction of the Plate & I introduce the volar locking plate along the distal radius volar cortex without screws. \\
Modification of Plate Positioning & I remove and reposition the volar plate to optimize alignment with the volar cortex. \\
Proximal Fixation & I insert cortical screws through the proximal plate holes into the intact diaphyseal shaft. \\
Verification & I perform fluoroscopic control with anteroposterior and lateral views to confirm reduction and screw placement. \\
\bottomrule
\end{tabular}
\end{table*}

\subsection{Replication of \baseline{}$_{(\beta)}$}
\label{app:downstream}
To validate our reimplementation of \baseline{}$_{(\beta)}^{*}$, we reproduce the zero-shot evaluation protocol of Perez~\etal~\cite{perez2025surglavi} (their Table~2) on the six benchmarks they use for phase and step recognition: Cholec80~\cite{twinanda2017endonet} and HeiChole~\cite{wagner2023heichole} (cholecystectomy), AutoLaparo~\cite{wang2022autolaparo} (hysterectomy), StrasbyPass70 and BernBypass70~\cite{lavanchy2024multibypass} (gastric bypass), and GraSP~\cite{ayobi2025grasp} (prostatectomy). Cholec80, AutoLaparo, and HeiChole provide only phase annotations and therefore cannot be evaluated on the cross-level tasks of \bench{}, which is why they are reported in this supplementary rather than in the main paper. Step recognition is reported only for GraSP, the only dataset for which SurgLaVi reports step-level results. \Cref{tab:downstream_full} includes the original \baseline{}$_{(\beta)}$ results reported by Perez~\etal{} alongside our reimplementation \baseline{}$_{(\beta)}^{*}$, allowing readers to assess replication fidelity.

\begin{table}[ht]
\centering
\caption{\textbf{Reimplementation validation.} Zero-shot downstream performance (video-averaged, \%) comparing the original \baseline{}$_{(\beta)}$ results from Perez~\etal~\cite{perez2025surglavi} (their Table~2) with our reimplementation \baseline{}$_{(\beta)}^{*}$, trained on the same data with the same procedure.
}
\label{tab:downstream_full}
\setlength{\tabcolsep}{2.5pt}
\small
\begin{tabular}{@{}lrr@{\hskip 8pt}rr@{}}
\toprule
& \multicolumn{2}{c}{\textbf{Phase ($\uparrow$)}} & \multicolumn{2}{c}{\textbf{Step ($\uparrow$)}} \\
\cmidrule(lr){2-3}\cmidrule(lr){4-5}
& \textit{Accuracy} & \textit{F1} & \textit{Accuracy} & \textit{F1} \\
\midrule
\multicolumn{5}{@{}>{\columncolor{gray!15}[0pt][0pt]}l@{}}{\textit{Cholec80}} \\
\baseline{}$_{(\beta)}$ & 58.0 & 39.4 & --- & --- \\
\baseline{}$_{(\beta)}^{*}$ & 57.8 & 34.5 & --- & --- \\
\multicolumn{5}{@{}>{\columncolor{gray!15}[0pt][0pt]}l@{}}{\textit{AutoLaparo}} \\
\baseline{}$_{(\beta)}$ & 55.7 & 46.0 & --- & --- \\
\baseline{}$_{(\beta)}^{*}$ & 58.0 & 44.3 & --- & --- \\
\multicolumn{5}{@{}>{\columncolor{gray!15}[0pt][0pt]}l@{}}{\textit{HeiChole}} \\
\baseline{}$_{(\beta)}$ & 57.0 & 44.0 & --- & --- \\
\baseline{}$_{(\beta)}^{*}$ & 63.6 & 35.5 & --- & --- \\
\multicolumn{5}{@{}>{\columncolor{gray!15}[0pt][0pt]}l@{}}{\textit{BernBypass70}} \\
\baseline{}$_{(\beta)}$ & 18.3 & 15.1 & --- & --- \\
\baseline{}$_{(\beta)}^{*}$ & 23.9 & 17.4 & --- & --- \\
\multicolumn{5}{@{}>{\columncolor{gray!15}[0pt][0pt]}l@{}}{\textit{StrasbyPass70}} \\
\baseline{}$_{(\beta)}$ & 31.2 & 26.1 & --- & --- \\
\baseline{}$_{(\beta)}^{*}$ & 29.5 & 22.7 & --- & --- \\
\multicolumn{5}{@{}>{\columncolor{gray!15}[0pt][0pt]}l@{}}{\textit{GraSP}} \\
\baseline{}$_{(\beta)}$ & 34.8 & 28.0 & 14.2 & 11.1 \\
\baseline{}$_{(\beta)}^{*}$ & 25.1 & 17.7 & 17.9 & 9.0 \\
\bottomrule
\end{tabular}
\end{table}

\subsection{Additional Training Details}
\label{app:training}

\noindent\textbf{Per-level training dynamics.}
The contrastive loss converges at different rates per hierarchy level: phase loss (0.032) $<$ step loss (0.050) $<$ action loss (0.109) at the end of training.
Phases (Level~3), whose rich captions map to few unique labels, are easiest to discriminate contrastively.
Actions (Level~1), being brief action-level descriptions, are hardest.
\ours{} preserves this natural difficulty gradient across hierarchy levels rather than collapsing it.

\noindent\textbf{Optimizer.} AdamW with $\text{lr}{=}10^{-4}$, weight decay $0.02$, and gradient clipping at max norm $1.0$.
LayerNorm parameters, bias terms, positional embeddings, and learnable scalar parameters (temperature, curvature, projection scales) are excluded from weight decay.

\noindent\textbf{Scheduler.} OneCycleLR with 5-epoch linear warmup (initial lr $= 10^{-6}$), cosine annealing, and a final learning rate of $10^{-6}$.

\noindent\textbf{Batch Composition.} Each GPU processes 39 samples: 30 randomly sampled and 9 from 3 forced triplets (one caption per hierarchy level from the same video).
Total effective batch size across 8 GPUs: 312.

\noindent\textbf{Numerical Stability.} All Lorentz distance computations (including $\acosh$) are performed in float32, even when the rest of the forward pass uses bfloat16 mixed precision. The argument to $\acosh$ is clamped to $[1+\epsilon, \infty)$ to prevent NaN gradients.

\noindent\textbf{Learnable Parameters.} Temperature $\tau$ and curvature $c$ are stored in log-space and clamped to $[0.01, 1.0]$ and $[0.1, 10.0]$ respectively.
Projection scaling factors $\alpha_v, \alpha_t$ are stored in log-space, initialized to $\log(1/\sqrt{256})$, and clamped to $(-\infty, 0]$ so that the effective scales $e^{\alpha_v}, e^{\alpha_t} \in (0, 1]$. At convergence: $\tau{=}0.035$, $c{=}0.815$, $e^{\alpha_v}{=}0.054$, $e^{\alpha_t}{=}0.060$.

\section{\ours{} Ablation}
\label{app:ablation}
\begin{table}[t!]
\centering
\caption{\textbf{Ablation of \ours{} components} (video-averaged accuracy, \%). All variants use Lorentz distance. The full model achieves the best overall performance, with both components contributing complementary gains. \textbf{Bold}\,=\,best per dataset.}
\label{tab:ablation}
\setlength{\tabcolsep}{3pt}
\small
\begin{tabular}{@{}lrrr@{}}
\toprule
& \textbf{Phase ($\uparrow$)} & \textbf{Step ($\uparrow$)} & \textbf{Multi-level ($\uparrow$)} \\
\midrule
\multicolumn{4}{@{}>{\columncolor{gray!15}[0pt][0pt]}l@{}}{\textit{MIPO}} \\
Lorentz only            & 13.8 & 11.9 & 12.8 \\
+ Entailment cones      & 9.3  & 20.6 & 13.8 \\
+ Triplet sampler       & 16.8 & \textbf{22.4} & \textbf{19.4} \\
+ Both (\ours{})        & \textbf{20.9} & 17.6 & 19.2 \\
\multicolumn{4}{@{}>{\columncolor{gray!15}[0pt][0pt]}l@{}}{\textit{StrasbyPass70}} \\
Lorentz only            & 20.4 & 24.1 & 22.2 \\
+ Entailment cones      & 24.3 & 20.2 & 22.2 \\
+ Triplet sampler       & 23.5 & 25.4 & 24.4 \\
+ Both (\ours{})        & \textbf{30.2} & \textbf{32.0} & \textbf{31.1} \\
\multicolumn{4}{@{}>{\columncolor{gray!15}[0pt][0pt]}l@{}}{\textit{BernBypass70}} \\
Lorentz only            & 12.6 & 17.7 & 14.9 \\
+ Entailment cones      & 13.8 & 17.4 & 15.5 \\
+ Triplet sampler       & 13.0 & 21.1 & 16.6 \\
+ Both (\ours{})        & \textbf{17.4} & \textbf{29.7} & \textbf{22.7} \\
\multicolumn{4}{@{}>{\columncolor{gray!15}[0pt][0pt]}l@{}}{\textit{GraSP}} \\
Lorentz only            & 24.2 & 12.9 & 17.7 \\
+ Entailment cones      & 26.7 & \textbf{22.0} & \textbf{24.2} \\
+ Triplet sampler       & \textbf{28.1} & 13.8 & 19.7 \\
+ Both (\ours{})        & 26.6 & 14.7 & 19.8 \\
\midrule
\multicolumn{4}{@{}>{\columncolor{cyan!4}[0pt][0pt]}l@{}}{\textbf{\textit{Overall}}} \\
Lorentz only            & 17.8 & 16.7 & 16.9 \\
+ Entailment cones      & 18.5 & 20.1 & 18.9 \\
+ Triplet sampler       & 20.4 & 20.7 & 20.0 \\
+ Both (\ours{})        & \textbf{23.8} & \textbf{23.5} & \textbf{23.2} \\
\bottomrule
\end{tabular}
\end{table}

We ablate the three components of \ours{} --- Lorentz distance, hierarchy-aware triplet batch sampler, and entailment cones --- by training variants with different combinations.
\Cref{tab:ablation} compares four configurations: (i)~Lorentz distance only (no batch sampler, no cones), (ii)~Lorentz distance with entailment cones but random batches, (iii)~Lorentz distance with the triplet batch sampler but no cones, and (iv)~the full \ours{} with all three components.

Both components improve over the Lorentz-only baseline, but they play different roles. The triplet batch sampler is more reliable, improving both phase and step on every multi-level dataset. Where the full model pulls ahead is on the Bypass datasets, where cones and the sampler reinforce each other and produce gains of up to +6.7,pp on phase and +8.6,pp on step over the sampler alone. The exceptions follow the expected pattern. On GraSP, whose DAG-like hierarchy violates the tree assumption behind entailment cones, adding cones provides no further benefit at the phase level. On MIPO, the sampler alone edges out the full model on step accuracy, though cones still improve phase accuracy (+4.1,pp). Overall, the full model improves over the Lorentz-only baseline by +6.0,pp on phase and +6.9,pp on step, raising the multi-level score from 16.9\% to 23.2\%.

\clearpage
\noindent\textbf{Sensitivity to $\lambda$.}
The entailment loss weight $\lambda$ (\cref{eq:total}) controls how strongly the model enforces parent--child containment via entailment cones.
\Cref{tab:lambda} reports phase and step accuracy of \ours{} across the four multi-level datasets for five values of $\lambda$, with all other hyperparameters held fixed.

No single value of $\lambda$ dominates across all datasets. While a lower value ($\lambda{=}0.05$) favor phase accuracy, a higher value ($\lambda{=}0.3$) favor step accuracy, $\lambda{=}0.1$ strikes the best balance between both levels, achieving the highest overall multi-level score (23.2\%) and the best or near-best results on three of four datasets. 

\begin{table}[t!]
  \centering
  \caption{\textbf{Sensitivity of \ours{} to the entailment loss weight $\lambda$} (video-averaged accuracy, \%). All variants use the full
  configuration (Lorentz + cones + triplet sampler), only $\lambda$ varies. We select $\lambda{=}0.1$ as it yields the most balanced performance across both levels. \textbf{Bold}\,=\,best per dataset.}
  \label{tab:lambda}
  \setlength{\tabcolsep}{3pt}
  \small
  \begin{tabular}{@{}lrrr@{}}
  \toprule
  & \textbf{Phase ($\uparrow$)} & \textbf{Step ($\uparrow$)} & \textbf{Multi-level ($\uparrow$)} \\
  \midrule
  \multicolumn{4}{@{}>{\columncolor{gray!15}[0pt][0pt]}l@{}}{\textit{MIPO}} \\
  $\lambda{=}0.01$ & 18.6 & 17.4 & 18.0 \\
  $\lambda{=}0.05$ & \textbf{22.8} & 20.0 & \textbf{21.4} \\
  $\lambda{=}0.1$  & 20.9 & 17.6 & 19.2 \\
  $\lambda{=}0.2$  & 17.5 & \textbf{23.6} & 20.3 \\
  $\lambda{=}0.3$  & 13.7 & 19.9 & 16.5 \\
  \multicolumn{4}{@{}>{\columncolor{gray!15}[0pt][0pt]}l@{}}{\textit{StrasbyPass70}} \\
  $\lambda{=}0.01$ & 22.1 & 26.5 & 24.2 \\
  $\lambda{=}0.05$ & 27.2 & 31.1 & 29.1 \\
  $\lambda{=}0.1$  & \textbf{30.2} & 32.0 & \textbf{31.1} \\
  $\lambda{=}0.2$  & 26.1 & 25.4 & 25.7 \\
  $\lambda{=}0.3$  & 28.8 & \textbf{32.3} & 30.5 \\
  \multicolumn{4}{@{}>{\columncolor{gray!15}[0pt][0pt]}l@{}}{\textit{BernBypass70}} \\
  $\lambda{=}0.01$ & 10.2 & 22.8 & 15.2 \\
  $\lambda{=}0.05$ & \textbf{18.2} & 26.3 & 21.9 \\
  $\lambda{=}0.1$  & 17.4 & \textbf{29.7} & \textbf{22.7} \\
  $\lambda{=}0.2$  & 17.7 & 23.4 & 20.4 \\
  $\lambda{=}0.3$  & 17.6 & 28.6 & 22.4 \\
  \multicolumn{4}{@{}>{\columncolor{gray!15}[0pt][0pt]}l@{}}{\textit{GraSP}} \\
  $\lambda{=}0.01$ & 31.1 & \textbf{18.3} & 23.9 \\
  $\lambda{=}0.05$ & 33.8 & 10.0 & 18.4 \\
  $\lambda{=}0.1$  & 26.6 & 14.7 & 19.8 \\
  $\lambda{=}0.2$  & \textbf{35.7} & 17.7 & \textbf{25.1} \\
  $\lambda{=}0.3$  & 28.6 & 16.1 & 21.5 \\
  \midrule
  \multicolumn{4}{@{}>{\columncolor{cyan!4}[0pt][0pt]}l@{}}{\textbf{\textit{Overall}}} \\
  $\lambda{=}0.01$ & 20.5 & 21.3 & 20.3 \\
  $\lambda{=}0.05$ & \textbf{25.5} & 21.9 & 22.7 \\
  $\lambda{=}0.1$  & 23.8 & 23.5 & \textbf{23.2} \\
  $\lambda{=}0.2$  & 24.3 & 22.5 & 22.9 \\
  $\lambda{=}0.3$  & 22.7 & \textbf{24.2} & 22.2 \\
  \bottomrule
  \end{tabular}
\end{table}

\section{Qualitative Results}
\label{app:qualitative}

\begin{figure*}[th!]
    \centering
    \includegraphics[width=0.85\linewidth]{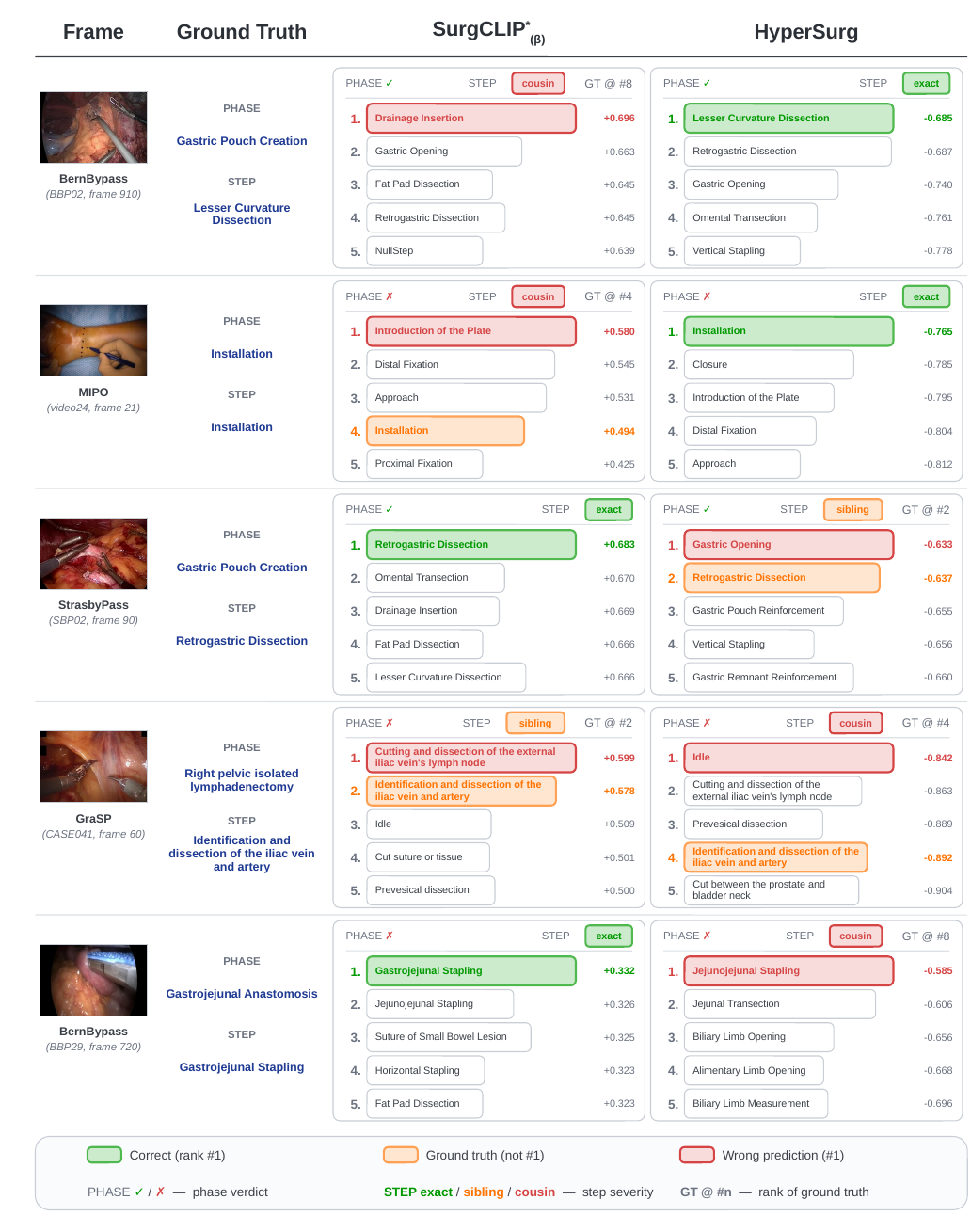}
    \caption{\textbf{Where does each model get it right, and when wrong, how bad is the error?} Each row shows a test frame with ground-truth labels and top-5 step predictions. 
    \textit{Top two rows:} \ours{} predicts the exact step while \baseline{}$_{(\beta)}^{*}$ predicts a cousin from a different phase. \textit{Row 3:} \ours{} predicts the wrong step but stays within the correct phase (sibling, 0.004 margin), while \baseline{}$_{(\beta)}^{*}$ gets the exact step. Flat metrics treat both errors identically. \textit{Row 4:} \ours{} predicts Idle, a cross-phase error on GraSP, consistent with its DAG-like hierarchy. \textit{Row 5 (subtree lock-in):} \ours{} commits to the wrong phase subtree, with all five predictions from Jejunojejunal Anastomosis instead of the ground-truth Gastrojejunal Anastomosis, while \baseline{}$_{(\beta)}^{*}$ ranks the correct step first. Both the strengths and this failure mode are consistent with the same geometric property: hierarchical containment helps when the correct subtree is identified, but can trap the model when sibling phases are visually similar.}

\label{fig:temporal_crosslevel_1}
\end{figure*}

To complement the quantitative results, \cref{fig:temporal_crosslevel_1} presents qualitative examples of both successful and failed predictions from \baseline{}$_{(\beta)}^{*}$ and \ours{} across all four hierarchical datasets.

%----------------
%Succ and failure examples

\end{document}